\documentclass[letterpaper]{article} 
\usepackage{aaai2027}
\nocopyright  
\usepackage[hyphens]{url}  
\usepackage{graphicx} 
\usepackage{natbib}  
\usepackage{caption} 
\usepackage{booktabs}
\usepackage{colortbl,xcolor}
\usepackage{xspace}
\usepackage{amssymb}
\usepackage{pdfpages}
\newcommand{\bench}{\textsc{AgentFootprint}\xspace}
\newcommand{\Stotal}{\ensuremath{S_{\mathrm{total}}}\xspace}
\newcommand{\Dup}{\ensuremath{D}\xspace}
\newcommand{\Alpha}{\ensuremath{\alpha}\xspace}
\newcommand{\Comp}{\ensuremath{C}\xspace}
\newcommand{\Rep}{\ensuremath{R}\xspace}
\definecolor{hl}{RGB}{235,242,250}
\title{The Hidden Footprint: Making Storage a First-Class Metric for
  LLM Agent Evaluation}
\author{Chenglin Yu\textsuperscript{\rm 1},
  Hongquan Gui\textsuperscript{\rm 1},
  Ying Yu\textsuperscript{\rm 2},
  Tao Zeng\textsuperscript{\rm 4},
  Ming Li\textsuperscript{\rm 1,3}\thanks{Corresponding author.}}
\affiliations{
  \textsuperscript{\rm 1}Department of Industrial and Systems
  Engineering, The Hong Kong Polytechnic University\\
  \textsuperscript{\rm 2}College of Economics and Management,
  Zhejiang Normal University\\
  \textsuperscript{\rm 3}Research Institute for Generative AI,
  The Hong Kong Polytechnic University\\
  \textsuperscript{\rm 4}Shanghai Juepei Flexible Supply Chain
  Technology Co., Ltd.\\
  chenglin.yu@polyu.edu.hk, ming.li@polyu.edu.hk
}

\begin{document}
\maketitle

\begin{abstract}
LLM agent benchmarks measure task completion, reliability, and
inference cost, but not the persistent data an agent run leaves on
disk, including logs, context snapshots, checkpoints, and debug traces. These
bytes are absent from the leaderboards we survey, yet bear on
whether an agent system can be deployed on desktops, under
data-retention compliance regimes, or at fleet scale. To our knowledge,
\bench is the first systematic cross-framework benchmark of post-run
agent storage footprint. Its serialization-aware metric suite covers
total retention, channel composition, duplication, growth exponent,
compressibility, and a conversation-history reconstructability score. It addresses a
measurement trap: na\"ive byte-level measurement understates
duplication by an order of magnitude, because database paging and
JSON escaping obscure repeated content. The footprint has two
determinants: the logical volume generated by the agent's behavior
and the amplification added by the persistence layer. A fixed-trace
control isolates the latter: the same
trajectory replayed through each persisting framework's adapter
yields retained sizes differing by $6.7\times$. Among configurations
achieving 100\% accuracy under identical models, tools, and tasks,
retained bytes differ by $15.7\times$. The defaults bundle different recovery and audit
capabilities, so the suite reports storage jointly with accuracy and
reconstructability. Three full-history configurations grow
superlinearly on a repeated-observation stress task, and on a
deliberately minimal write task framework residue exceeds the
delivered output files by orders of magnitude. Exported trajectories
from 108 instance-normalized SWE-bench Verified submissions span
three orders of magnitude in per-instance volume with no detectable
correlation with resolve rate. A content-addressed store reduces
retention $4.8$--$32.7\times$ while preserving all properties
checked by our restoration tests, including every reconstructability
score. Exact history reconstruction does not require megabytes, and
these storage metrics can be reported alongside inference cost.
\end{abstract}

\section{Introduction}
\label{sec:intro}

LLM agents---systems that pursue multi-step tasks by interleaving model
calls with tool use---are moving from demos into production, and the
community evaluates them along an expanding set of axes: task success on
GAIA~\citep{gaia} and SWE-bench~\citep{swebench}, reliability via
pass\textasciicircum{}k on $\tau$-bench~\citep{taubench}, and, following
\citet{aiagentsthatmatter}, inference cost. A recent survey of agent
evaluation catalogs more than a dozen such
metrics~\citep{kddsurvey}. None includes the \emph{storage footprint},
defined here as the bytes an agent execution leaves behind on disk.

Unlike inference cost, which is consumed when execution ends,
retention is a stock that persists and \emph{accumulates} across
runs. This persistence is simultaneously useful and costly. It is
the substrate for replay debugging, recovery, and compliance audits,
but it also enlarges long-term storage, governance, and deletion
obligations. The magnitudes involved are easy to underestimate. In a
production
deployment that we measured directly (a supply-chain quoting system;
Supp.\ App.~P), a single task retained 131\,MB, including 79\,MB
of framework residue, compared with 2.6\,MB of delivered artifacts.
Per-task retention had a median of 715\,KB and a 90th percentile of
81\,MB. The deployment as a whole has accumulated
${\sim}56$\,GB over ${\sim}10^3$ recorded rounds
(operator-attested). For illustration, at the same per-round average
a fleet processing $10^4$ rounds/day would accumulate
${\sim}560$\,GB daily if nothing is deleted.

Our goal is to make this axis measurable and comparable across agent
frameworks. Doing so is challenging for three reasons. First,
\emph{attribution}: an agent run scatters bytes across working
directories, session databases, hidden home-directory state, and
tokenizer caches. Deciding which bytes are task-attributable requires
per-run isolation, not a disk-usage command. Second,
\emph{serialization masking}: the same file content is re-stored many
times inside SQLite pages and JSON-escaped message arrays, where page
fragmentation and escape sequences defeat byte-level fingerprinting.
Raw-file chunking can report $\Dup{=}1.01$ for a store logically holding
the same content twelve times over. Third,
\emph{semantics}: raw byte counts alone cannot distinguish an agent that
retains much to support auditing from one that retains much because of
avoidable storage amplification. The metric suite must indicate
which recovery and audit capabilities the retained bytes support.

We present \bench, a benchmark and measurement harness for these
challenges. Each (framework, task) pair runs in a fresh sandbox whose
workspace and home delta is captured and attributed (challenge one). Our
meter analyzes \emph{logical content streams} (SQLite cells and JSONL
records) rather than raw files, and injects content probes that are
matched under raw and JSON-escaped encodings (challenge
two). On top of this substrate we define six measurements: total retention
\Stotal, a descriptive storage-channel composition, duplication factor
\Dup, growth exponent \Alpha fitted over controllable-horizon tasks,
long-window compressibility \Comp, and a 0--3 history-reconstructability
score \Rep asking whether retained bytes reconstruct the conversation
history behind step $k$ (challenge three). Throughout, storage is a system-level
\emph{resource} dimension, read jointly with accuracy and \Rep,
never a capability by itself. Its two determinants are separated by
a fixed-trace control (\S\ref{sec:results:fixedtrace}).

Measuring eight representative frameworks (LangGraph, AutoGen, CrewAI,
SmolAgents, OpenAI Agents SDK, LlamaIndex, Agno, and
InfiAgent) under identical backends, tools, and tasks
yields three findings. (i)~\emph{Equal-accuracy dispersion}: at
among configurations achieving 100\% task accuracy, retained bytes differ by $15.7\times$
(LangGraph vs.\ CrewAI), and one framework retains nothing at
all and accordingly can recover or resume nothing. (ii)~\emph{Growth regimes}: frameworks that
persist full history at every step exhibit \Alpha up to 1.95: a 200-round
monitoring loop over one 2\,KB status file leaves 323\,MB, while
frameworks with windowed context management stay near-linear. (iii)~\emph{Residue dominates artifacts}: when agents produce
legitimate output files, framework residue exceeds those artifacts by
up to $24{,}033\times$.

Public agent submissions exhibit an analogous dispersion in exported
trajectory volume: across 108
instance-normalized SWE-bench Verified submissions, per-instance volume spans
$1{,}617\times$ with no detectable correlation with resolve rate. A
content-addressed store removes most retained bytes while all restore
checks pass and every \Rep score is reproduced.

Our contributions are:
\begin{itemize}
  \item \textbf{A metric suite and measurement methodology} for agent
  storage footprint (\Stotal, composition, \Dup, \Alpha, \Comp, \Rep),
  including the finding that serialization masks duplication from na\"ive
  measurement by up to $12\times$ (\S\ref{sec:metrics}).
  \item \textbf{\bench}, a reproducible harness and task suite measuring
  eight frameworks over 1{,}061 sandboxed runs: a 509-run main study
  (three repetitions plus ablation, growth, write, exploratory,
  heterogeneous, and fixed-trace suites) plus 552 replication and
  variant runs (two further backends, growth seeds, executed
  history-only forms, shared threads, replay-source tasks), with
  container-validated
  isolation (\S\ref{sec:bench}--\S\ref{sec:results}; full accounting
  in Supp.\ App.~C).
  \item \textbf{An in-the-wild study} harvesting all 134 SWE-bench
  Verified submissions (108 instance-normalized), finding a $1{,}617\times$ volume
  spread with no detectable correlation between exported volume and success
  ($\tau_b=0.080$, $p=0.222$) (\S\ref{sec:wild}).
  \item \textbf{A headroom demonstration}: an existence proof, not a
  proposed storage system. A content-addressed compactor shows that the
  measured reconstructability score survives at $4.8$--$32.7\times$ fewer
  bytes than the frameworks' defaults (\S\ref{sec:cas}).
\end{itemize}

\section{Related Work}
\label{sec:related}

\bench expands \emph{what} agent benchmarks measure rather than
\emph{how well} agents score. \citet{aiagentsthatmatter}
introduced inference cost as a mandatory reporting axis and showed simple
baselines dominate the cost--accuracy Pareto frontier. $\tau$-bench added
reliability via pass\textasciicircum{}k~\citep{taubench}. Storage footprint
is orthogonal to both and differs in kind: cost is a flow that stops
with the task. Retention is a stock that persists, compounds, and
carries compliance and reproducibility obligations.
Efficient Agents studies effectiveness against inference
cost~\citep{efficientagents}. MAFBench reports architecture-dependent latency
and coordination costs~\citep{mafbench}. Neither measures persistent residue, which \bench isolates.

A second line optimizes what agents keep \emph{in context or in memory
stores}. Memory systems (MemRefine~\citep{memrefine}) compress memory banks
for retrieval. Context compaction (Slipstream~\citep{slipstream})
shortens prompts. Both target
the model-facing representation. The persistence layer can still
checkpoint every intermediate underneath (one measured
framework does). We measure the \emph{system-wide residue}, coupled to
context policy through \Alpha.

Trajectory platforms (LangSmith, Langfuse, AgentOps) persist agent
runs as a product feature but publish no efficiency
measures (potential reporters of such metrics, not sources).
CLP~\citep{clp} and zstd~\citep{zstd} compress syntactically (zstd is
our baseline \Comp). Our semantic metrics capture what compression
neither reveals nor repairs: what is duplicated, how retention
scales, whether bytes reconstruct history.

Finally, the gap is explicit in the surveys: \citet{kddsurvey}'s
metric catalog spans completion, step-efficiency, tool correctness,
plan quality, latency, and cost, but no storage dimension. GAIA,
SWE-bench, and $\tau$-bench report none. To our knowledge, \bench is the first systematic cross-framework
benchmark of post-run storage footprint for LLM agents. Adjacent lines (workflow
provenance~\citep{davidson2008provenance,herschel2017provenance},
LSM storage amplification~\citep{oneil1996lsm}) instrument different
objects. We import their techniques as mitigations.

\section{Storage Footprint Metrics}
\label{sec:metrics}

\begin{figure*}[t]
\centering
\includegraphics[width=0.98\textwidth,page=1]{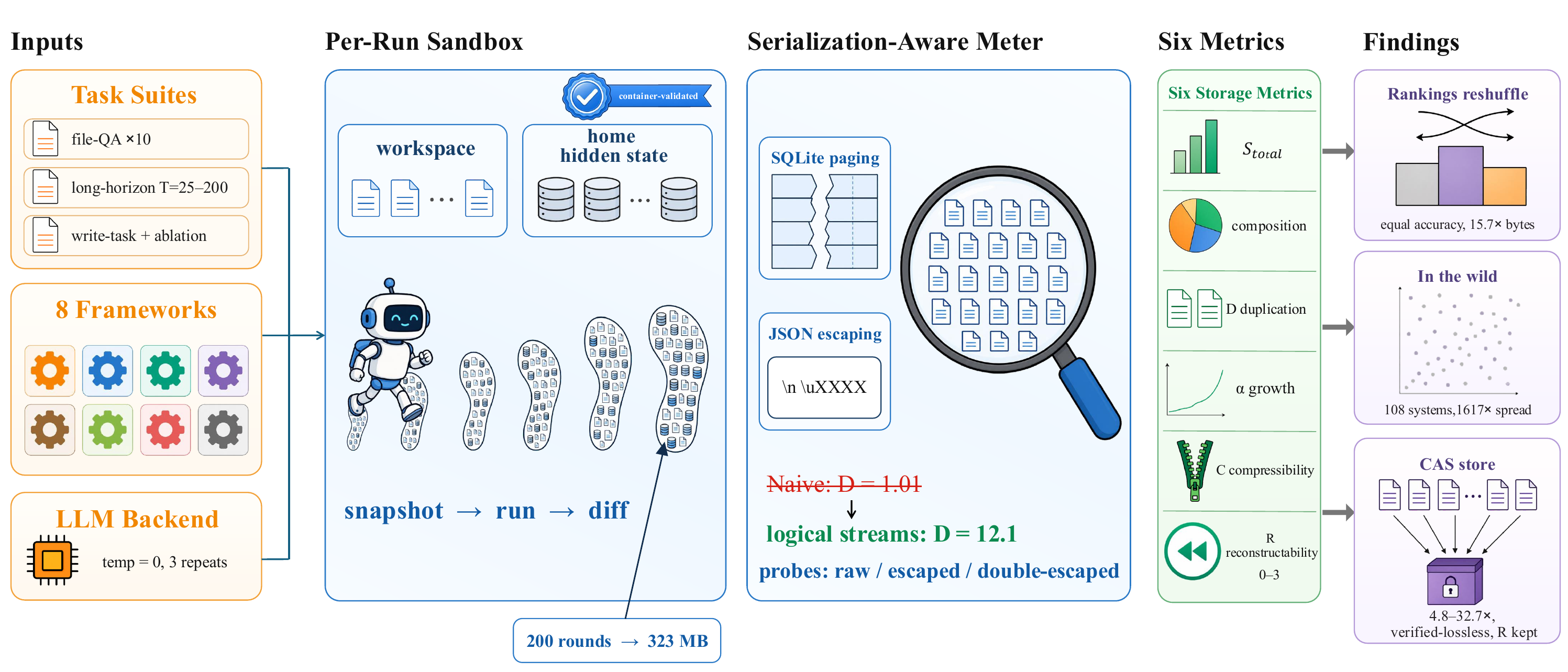}
\caption{\bench pipeline: each (framework, task, repetition) runs in a
fresh sandbox. The filesystem delta is analyzed as \emph{logical
content streams} rather than raw bytes, yielding the six metrics in
\S\ref{sec:metrics} that drive
\S\ref{sec:results}--\S\ref{sec:cas}.}
\label{fig:pipeline}
\end{figure*}

\subsection{What Counts: Boundary and Taxonomy}
\label{sec:metrics:boundary}

We define the storage footprint of an agent execution---post-run
local retention within the declared sandbox boundary---as all bytes
present after the run, attributable to it, absent before it: files
created or grown in the task workspace and every framework-side
store (session/checkpoint databases, logs, snapshots, and traces), including
hidden home-directory state. \emph{Footprint} and \emph{retention} are used interchangeably;
\emph{residue} is the framework-side portion (all but agent-produced
workspace artifacts). Environment artifacts shared across tasks (model weights,
tokenizer/package caches) are excluded, although their volume is
reported for auditability. Measurement uses a per-run sandbox: a fresh workspace
and home whose file inventory is snapshotted before and diffed after
the run (\S\ref{sec:bench:harness}); Supp.\ App.~N audits this
delta accounting over all 509 main-study sandboxes.

Retained bytes are not equally justified. We classify them as
\emph{decisions} (model outputs, generally not regenerable),
\emph{observations} (tool and environment returns, some of which are
likewise unrecoverable),
\emph{derivations} (framework-generated intermediates: snapshots,
summaries, checkpoints), and \emph{duplicates} (the second through
$n$-th appearance of the same bytes), separating what must be
kept, what may be referenced, and what is overhead. The taxonomy is conceptual; the benchmark reports
deterministic storage-channel and duplication measurements, not
automated semantic attribution of every byte.

\subsection{Six Metrics}
\label{sec:metrics:defs}

On this foundation we define six measurements per (framework, task) run:

\begin{description}
  \item[\Stotal] Retained bytes (apparent-length filesystem delta;
  Supp.\ App.~N), divided into workspace artifacts and
  framework-side residue.
  \item[Store composition] \Stotal grouped into state/checkpoints,
  debug/event logs, conversation/actions, and other channels by
  deterministic, adapter-audited path rules.
  \item[\Dup] Duplication factor $\Stotal^{\mathrm{logical}} /
  S_{\mathrm{unique}}$. Here, $S_{\mathrm{unique}}$ sums unique
  content-defined chunks (FastCDC~\citep{fastcdc} with SHA-256) over
  \emph{logical streams}. Both live in logical content bytes (SQLite
  headers, indexes, and free pages enter neither); $\Dup{=}2$ means
  the average content byte is stored twice; an exact-match lower
  bound under compressed or delta-encoded representations
  (Supp.\ App.~K); physical-layout overhead is \Stotal{} vs.\ the
  logical total.
  \item[\Alpha] Growth exponent from fitting $\log \Stotal \sim
  \Alpha \log T$ at $T\in\{25,50,100,200\}$ on the
  repeated-observation stress task---a descriptive statistic;
  per-horizon retention is primary (Supp.\ App.~G).
  \item[\Comp] Compressibility: $\Stotal$ divided by its size under
  zstd~-19 with a 128\,MB long-range window over the concatenated
  retained files. This is a compression bound, not an archive. We
  report it to separate syntactic compression gains from semantic structure.
  \item[\Rep] Conversation-history reconstructability, 0--3: given call
  boundaries (\S\ref{sec:results:replay}), can the retained bytes reconstruct
  the conversation history the model received at step $k$? $\Rep{=}0$: nothing retained;
  $\Rep{=}1$: bytes but no per-call structure; $\Rep{=}2$: per-call
  structure but no complete copy of the observations; $\Rep{=}3$: at least
  one channel holds a complete, exact record. Automated scoring yields
  a \emph{candidate} grade; reported grades additionally require the
  adapter's serialization contract to pass the field-by-field
  reconstruction of \S\ref{sec:results:replay} (construct check:
  Supp.\ App.~O).
\end{description}

\subsection{Measurement Methodology: Serialization Masks Duplication}
\label{sec:metrics:method}

A key methodological finding is that raw-file chunking substantially
understates duplication.
Running content-defined chunking over raw retained files reports
$\Dup=1.01$ for LangGraph's checkpoint store, although zstd compresses
it by $142\times$. This apparent discrepancy indicates missed logical
repetition. Two serialization mechanisms mask
the duplication. SQLite fragments large values across 4\,KB pages
interleaved with headers and pointers, so chunk boundaries never align
across copies; and JSON escaping rewrites the bytes of embedded content
(\texttt{\textbackslash n}, \texttt{\textbackslash uXXXX}), so
trajectory copies no longer match the source bytes. Our meter therefore extracts \emph{logical streams} before
fingerprinting: every SQLite cell and JSONL line is chunked as its own
stream, while other files are treated as whole streams. Under logical streams the same store
measures $\Dup=12.1$ (a single store; Table~\ref{tab:main}'s
30-run means are $\Dup$ 12.8, \Comp 136); the duplication was
present but obscured by
serialization, which may partly explain why this axis has previously
gone unmeasured.

Cross-representation copies still evade chunk matching, so we add
\emph{content probes}: fixed
substrings and whole lines sampled deterministically from every input
file, searched in all logical streams under raw and
JSON-escaped encodings; doubly-nested encodings are only partially
detected (Supp.\ App.~K). The mean occurrence count (\emph{echo}) estimates
how many times the input content is stored; the same probes power the
\Rep score's completeness check, and whole-line probes survive
re-formatting such as line-numbered tool outputs.
Supp.\ App.~K calibrates \Dup and echo on synthetic
stores with known duplication.

\section{The AgentFootprint Benchmark}
\label{sec:bench}

\subsection{Task Suites}
\label{sec:bench:tasks}

\bench uses six task families, each isolating one storage behavior,
plus a control. The \emph{file-QA suite} (10 tasks) gives the agent ten
$\sim$60\,KB documents (600\,KB total) and five sequential questions whose
answers are unique registry codes embedded in the files; questions
revisit files in an A,B,A,C,A pattern with instructed re-reading
(read amplification). Grading is exact substring match: no LLM
judge. The
\emph{long-horizon suite} fixes a 2\,KB status file re-read for
$T\in\{25,50,100,200\}$ rounds, isolating \Alpha from content
growth. The \emph{write-task suite} (5 tasks) reads
two documents and produces a summary file via a \texttt{write\_file}
tool (legitimate artifacts vs.\ residue). The
\emph{exploratory-retrieval suite} (5 tasks) removes every adversarial
element (\S\ref{sec:results:natural}). Two \emph{heterogeneous families} (five tasks
each), \emph{edit} and \emph{data analysis}, are
read--transform--write tasks graded \emph{jointly}: correct answer
\emph{and} artifact delivered in the shared workspace
(\S\ref{sec:results:het}). Finally,
\emph{ablation runs} re-execute the file-QA suite with persistence
disabled.

\subsection{Frameworks and Fairness Protocol}
\label{sec:bench:frameworks}

We measure eight representative frameworks: LangGraph
1.2.8~\citep{langgraph}, AutoGen 0.7.5~\citep{autogen}, CrewAI
1.15.2~\citep{crewai}, SmolAgents 1.26.0~\citep{smolagents}, OpenAI Agents
SDK 0.18.0~\citep{openaiagents}, LlamaIndex 0.14.23~\citep{llamaindex},
Agno 2.7.1~\citep{agno}, and InfiAgent 3.12.24~\citep{infiagent}. The roster
covers widely used open-source agent APIs with distinct persistence
mechanisms. Agno and InfiAgent provide two independent
implementations of windowed context management (a runtime forgetting
loop and a bounded per-request window), so the growth study
(\S\ref{sec:results:growth}) can compare full-history and
bounded-history policies.

Fairness comes from configuring each framework in its \emph{documented
durable-session configuration} (the persistence setup its own
documentation describes, selected by a predefined protocol), not from
tuning any of them for storage. Table~\ref{tab:main} lists the measured mechanisms. All frameworks receive identical task text, the same minimal tools
(list, read, write where applicable), the same backend model
(DeepSeek-V4-Flash, served by one provider, temperature~0), and three
independent repetitions. We measure the storage produced by the
documented default configuration. The exact adapter code ships in the artifact,
and \S\ref{sec:discussion} reports a configuration-sensitivity check
(AutoGen's save cadence).

This protocol measures the \emph{system-level default
footprint}: what the documented configuration leaves on disk. Its
limit: these documented configurations support
different capabilities (Supp.\ App.~R), so equal-\Rep comparisons
rank what defaults retain for the same reconstructability score,
not capability-normalized efficiency (\S\ref{sec:discussion}).

\subsection{Harness}
\label{sec:bench:harness}

The harness (Figure~\ref{fig:pipeline}) runs each (framework, task,
repetition) in a fresh sandbox. The workspace holds the task corpus,
and \texttt{HOME}/XDG paths are redirected into the sandbox to capture
hidden state. The adapter runs the framework's quickstart-style agent,
and the meter diffs both inventories. Re-running LangGraph (three
tasks) in per-run Docker containers changed measurements by
0--8.5\%, below the 11\% repetition-level standard deviation. Harness, generators, adapters, and per-run records ship in the
artifact. Supp.\ App.~C enumerates all 1{,}061 runs and states the
statistical units and aggregation conventions.

\section{Controlled Results}
\label{sec:results}

\subsection{Same Accuracy, $15.7\times$ the Bytes}
\label{sec:results:main}

Table~\ref{tab:main}(a) presents the file-QA suite (eight
frameworks, ten tasks, three repetitions): among the five persisting
configurations with 100\% task accuracy, documented defaults span $15.7\times$ in
retained bytes (bootstrap 95\% CI $[15.0,16.3]$; Supp.\ App.~C).
Across the six $\Rep{=}3$ systems, minimal history-retaining forms span
$3.9\times$: three executed variants, two native forms, and one
disclosed post-hoc split (\S\ref{sec:results:replay}; Supp.\ App.~Q).
Across the full roster, Agno answers 99.3\% and AutoGen 86\%; the other
six answer every question correctly. Persisting frameworks retain from 0.32\,MB
(CrewAI) to 5.10\,MB (LangGraph) per task on a 600\,KB corpus, and
SmolAgents retains exactly zero bytes.
Composition (Table~\ref{tab:main}b) shows retention concentrated in
different channels: checkpoints (LangGraph), event logs (AutoGen),
conversation stores and debug traces (InfiAgent; setup staging
excluded; including it raises its mean to 2.72\,MB, above AutoGen;
Supp.\ App.~N). The duplication factor separates
architectural families: full-history checkpoint/snapshot designs
(\Dup: LangGraph 12.8, AutoGen 8.4, LlamaIndex 4.2) are most redundant, while
windowed or append-only designs stay below 2.6. The echo metric makes
it concrete: probes from the same files, read equally often, appear
$8.1\times$ under LangGraph but only $0.5\times$ under OpenAI Agents.

\begin{table*}[t]
\centering
\begingroup
\footnotesize
\setlength{\tabcolsep}{4pt}
\begin{tabular*}{\textwidth}{@{\extracolsep{\fill}}llrrrrrrcr}
\toprule
Framework & Persistence mechanism & \Stotal $\downarrow$ & \Dup $\downarrow$ & Echo $\downarrow$
 & \Comp & \Alpha & Acc.\ $\uparrow$ & \Rep $\uparrow$ & CAS $\downarrow$ \\
\midrule
LangGraph      & SQLite checkpointer            & 5.10$\pm$0.58\,MB & 12.8 & 8.1$\times$ & 136 & 1.74$\pm$0.01 & 100\% & 3 & 0.50\,MB (10.3$\times$) \\
AutoGen        & per-turn state $+$ event log   & 2.49$\pm$0.33\,MB &  8.4 & 4.1$\times$ & 106 & 1.95$\pm$0.18 &  86\% & 3 & 0.07\,MB (32.7$\times$) \\
InfiAgent      & task stores $+$ raw traces     & 2.12$\pm$0.05\,MB &  1.8 & 1.7$\times$ &  43 & 1.15 & 100\% & 3 & 0.36\,MB (5.8$\times$)  \\
Agno           & SQLite session db              & 1.15$\pm$0.02\,MB &  2.5 & 0.8$\times$ &  42 & 0.98 & 99.3\% & 3 & 0.06\,MB (20.5$\times$) \\
LlamaIndex     & serialized \texttt{Context}    & 0.93$\pm{<}$0.01\,MB &  4.2 & 1.3$\times$ &  40 & 1.88$\pm$0.10 & 100\% & 3 & 0.09\,MB (10.3$\times$) \\
OpenAI Agents  & \texttt{SQLiteSession}         & 0.33$\pm$0.02\,MB &  1.6 & 0.5$\times$ &  14 & 0.73 & 100\% & 3 & 0.07\,MB (4.8$\times$)  \\
CrewAI         & task-output store              & 0.32$\pm{<}$0.01\,MB &  1.5 & 0.5$\times$ &  14 & 1.20 & 100\% & 1 & 0.05\,MB (6.1$\times$)  \\
\rowcolor{hl}
SmolAgents     & none (in-memory)               & 0\,B              &  --  & --          &  -- & --   & 100\% & 0 & --                      \\
\bottomrule
\end{tabular*}

\vspace{2pt}

\begin{tabular*}{\textwidth}{@{\extracolsep{\fill}}lrrrrrrrr}
\toprule
(b) Channel (MB/run) & LangGraph & AutoGen & InfiAgent & Agno & LlamaIndex & OpenAI Ag. & CrewAI & SmolAgents \\
\midrule
State/checkpoints    & 5.10 & 0.83 & 0.07 & 1.15 & 0.93 & 0.33 & 0.32 & -- \\
Debug/event logs     & --   & 1.65 & 0.86 & --   & --   & --   & --   & -- \\
Conversation/actions & --   & --   & 1.20 & --   & --   & --   & --   & -- \\
\midrule
Resume in fresh process & \checkmark & \checkmark & \checkmark & \checkmark & \checkmark & \checkmark & $\times$ & -- \\
\midrule
\Stotal, Qwen3.6-27B (MB) & 5.20 & 2.64 & 2.13 & 1.16 & 0.94 & 0.31 & 0.32 & 0 \\
\Stotal, Qwen2.5-7B (MB)  & 2.49 & 1.26 & 2.14 & 0.77 & 1.20 & 0.21 & 0.37 & 0 \\
\bottomrule
\end{tabular*}
\endgroup
\caption{Main results: the eight measured configurations as one
profile. \textbf{(a)}~File-QA suite (10 tasks $\times$ 3 repetitions;
\Stotal mean$\pm$SD over 30 runs; decomposition in
Supp.\ App.~C; spreads on unrounded values). Mechanism: documented
durable-session persistence (\S\ref{sec:bench:frameworks}); \Alpha:
growth exponent on the stress task (mean$\pm$SD over three seeds for
the full-history trio, single run otherwise;
\S\ref{sec:results:growth}, Supp.\ App.~G); \Rep: modal reconstructability score,
reconstruction-validated (\S\ref{sec:results:replay}); \Comp: zstd
long-range compressibility; CAS: post-CAS retention
(\S\ref{sec:cas}), first-repetition subset
(factor in parentheses). \textbf{(b)}~Channel composition of \Stotal,
process-boundary resume, and full-roster backend replication:
per-task \Stotal under two further backends, rank correlation with
the main study $\rho{=}0.96$ / $0.89$ (Supp.\ Apps.~B, F, J).}
\label{tab:main}
\end{table*}

\subsection{Growth Regimes under Repeated Observations}
\label{sec:results:growth}

\begin{figure}[t]
  \centering
  \includegraphics[width=\linewidth]{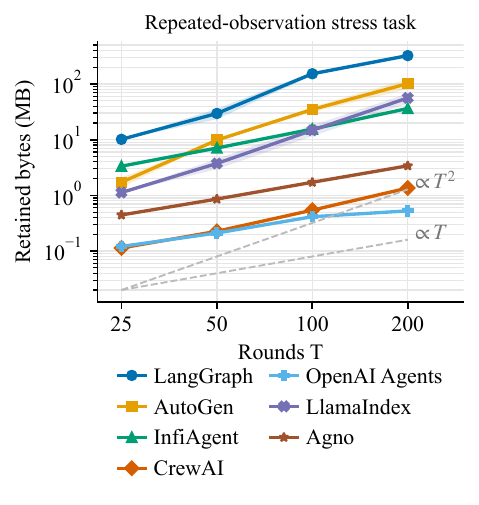}
    \caption{Retention growth on the repeated-observation stress task
  (log--log). Lines: means; bands: $\pm$1\,SD over three seeds
  (full-history trio). Full-range $\Alpha$ estimates are reported in
  Table~\ref{tab:main} and Supp.\ App.~G. Windowed and
  append-only designs remain linear or sublinear.}
  \label{fig:growth}
\end{figure}

On the controlled repeated-observation stress task, the retention
curves fall into three regimes (Figure~\ref{fig:growth};
\Alpha in Table~\ref{tab:main}; three seeded runs per horizon for the
full-history trio). Frameworks that persist the
full message history at every step (LlamaIndex, AutoGen, and
LangGraph; \Alpha 1.74--1.95) grow superlinearly:
monitoring one 2\,KB status file for 200 rounds leaves $323{\pm}26$\,MB
under LangGraph (\Dup ${\approx}$49), $102{\pm}21$\,MB under AutoGen. Frameworks
with windowed context (InfiAgent's forgetting loop) or bounded
per-round stores (CrewAI, Agno) stay near-linear (\Alpha
0.98--1.20): at $T{=}200$ (completed by every persisting framework)
they retain 1.4--36.4\,MB against LangGraph's $323{\pm}26$\,MB.

Accuracy stays high throughout (median 195 of 200 at $T{=}200$ across
persisting frameworks and seeds; minimum 171), so growth differences are
not task-failure artifacts. \Alpha is a workload-conditioned full-range fit: on
seed 1, local slopes over
$T{=}25{\to}100$ are 1.96--1.98 for all three heavy frameworks, but over
$100{\to}200$ they diverge (LlamaIndex 1.92, AutoGen 1.38, LangGraph
0.96) with no failed rounds and token volumes far below context
limits (per-seed spreads in Supp.\ App.~G). We suspect checkpoint-layer
consolidation. The repetitions show that variability concentrates in absolute volume
(LangGraph $T{=}100$: $153{\pm}10$\,MB) while the fitted exponent
remains stable
($1.74{\pm}0.01$). Four horizons at three runs each remain
insufficient to distinguish among candidate functional forms
(Supp.\ App.~G); the claim is
superlinear growth over the measured range---every seed of every
full-history framework fits $\Alpha{\geq}1.72$---not a scaling law.
Archival policy matters as much as architecture: LlamaIndex's
newest-snapshot-only form leaves 85--645\,KB
($\Alpha{\approx}0.98$); superlinear growth follows from the
\emph{append-all} archival policy, not from durable resumption
(Supp.\ App.~G).

On this fixed-content workload, read amplification and storage
amplification are separable, and the exponent tells them apart. OpenAI
Agents is sublinear
($\Alpha{=}0.73$): its session log appends each message once rather than
re-serializing the whole history, so the status file's content
accumulates linearly with reads ($\mathrm{echo}{=}120$ at
$T{=}200$) with no redundancy beyond that ($\Dup{=}1.0$). On this
workload the exponent therefore reflects context-retention policy:
whether a framework windows its history is observable
in how its disk usage scales. We do not claim the fitted exponent
transfers to arbitrary workloads.

\subsection{Persistence Ablations}
\label{sec:results:ablation}

Retention is not required for immediate task success on this short
retrieval suite: with persistence disabled, five frameworks
(LangGraph, AutoGen, LlamaIndex, OpenAI Agents, Agno; five-task
subset) keep their accuracy (100\% for four; AutoGen within run
variance: 22/25 vs.\ 19/25) and retain exactly zero bytes. Retention
supports replay, resumption, and recovery rather than immediate task
completion,
and the ablated configurations---like SmolAgents' default---provide
none ($\Rep{=}0$). InfiAgent exposes no debug-channel switch, so we
perform a post-hoc ablation by excluding its debug channel, which
accounts for 40\% of its footprint; removal \emph{raises} \Dup
1.75$\to$2.1. The ablation also
sharpens what the metrics measure: disabling persistence drives
\Stotal to zero and \Rep to 0 together, so the metric pair
distinguishes a configuration change from a structural property.

\subsection{Reconstructability}
\label{sec:results:replay}

Scoring one run per (framework, task) pair (80 runs), six frameworks
achieve $\Rep{=}3$ at retained sizes spanning $15\times$; CrewAI
retains bytes without per-call structure ($\Rep{=}1$), SmolAgents
nothing. InfiAgent's raw-trace channel supplies what its truncated
debug log lacks; multi-channel designs need channel-level scoring.
We validate the grade by \emph{full conversation-history
reconstruction}: each $\Rep{=}3$ framework re-ran three tasks behind
a logging proxy recording every request body (194 calls); we rebuilt
the conversation from the retained bytes and compared
field-by-field (roles, contents, tool-call ids/arguments, results,
order): with call boundaries taken from the proxy log, 194/194 calls
are exactly reconstructible as contiguous
subsequences (contracts in Supp.\ App.~I). System prompts are recoverable from retention in Agno and
InfiAgent; the rest keep them in code-side configuration (LangGraph
configures none).
A \emph{process-boundary resume probe}
(Supp.\ App.~J) confirms the grades operationally: every
$\Rep{=}3$ store resumes in a fresh process; CrewAI's
($\Rep{=}1$) does not.

Supp.\ App.~S plots default and post-CAS retention with \Rep and
accuracy; two observations follow. First, the same reconstructability score---$\Rep{=}3$---is
obtained at retained sizes from 0.33\,MB (OpenAI Agents) to 5.10\,MB
(LangGraph), and content addressing reduces all seven persisting
configurations to 0.05--0.50\,MB (0.06--0.50\,MB among the $\Rep{=}3$
configurations): most of the spread is removable redundancy, not a
requirement of the score. \Rep scores history reconstruction,
not every capability a store may serve (crash resume, workflow state,
debug provenance), so the spread is unexplained by reconstructability
alone, not established as pure waste (\S\ref{sec:discussion});
the executed minimal history-retaining forms
(\S\ref{sec:results:main}) converge at
0.31--0.33\,MB with accuracy and \Rep unchanged (Supp.\ App.~Q).
Second,
accuracy alone does not expose these storage differences: six frameworks
score 100\%, and Agno scores 99.3\%. A success-only leaderboard penalizes
AutoGen for its lower accuracy (86\%) but not for its 2.49\,MB footprint,
and it treats LangGraph as fully successful despite retaining
$15.7\times$ CrewAI's bytes. Section~\ref{sec:wild} examines the storage--success relationship
at scale.

\subsection{Beyond Adversarial Retrieval: Exploratory and
  Heterogeneous Tasks}
\label{sec:results:natural}

The stress suites deliberately provoke re-reads, so an
\emph{unprompted exploratory} check removes every adversarial
element: five multi-document tasks, free exploration via
\texttt{list\_files}. Dispersion persists (23/28 pairwise orderings
agree; persisting frameworks span $8.1\times$) and echo \emph{rises}
(LangGraph 8.1$\to$11.8$\times$); amplification is not an artifact
of adversarial design (AutoGen 47\% and SmolAgents 67\% differ in
accuracy and are excluded from equal-accuracy comparisons; Supp.\
App.~E).
\label{sec:results:het}

The two heterogeneous families (\S\ref{sec:bench:tasks}; all eight
frameworks, 80 runs; Supp.\ App.~D) grade success
\emph{jointly}: the answer must be correct \emph{and} the required
artifact delivered in the shared workspace, validated exactly against
the task specification.
Three patterns from the controlled suites reappear.
Equal-success dispersion: five configurations achieve full joint
success on both families (LangGraph, CrewAI, OpenAI Agents,
LlamaIndex, and Agno) and span $6.9\times$ in retention on edit tasks
(24--164\,KB) and $14.1\times$ on data tasks (8.8--124\,KB);
answer-only accuracy is broader (seven frameworks reach 100\% on edit)
but conflates delivered and undelivered work. The mechanism: LangGraph's checkpointer
echoes the input CSVs $7.8\times$ on data tasks; observation
amplification is not a file-QA artifact. The evaluation boundary:
InfiAgent answers every question correctly yet fails joint success on
all ten tasks, because it deposits the deliverable in its managed
home-side store rather than the shared workspace---exactly what the
workspace/home split of \S\ref{sec:metrics} makes visible; and
AutoGen's default tool-iteration budget again fails the
read--transform--write chain (0\% on both families).

\subsection{Fixed-Trace Control: Isolating the Persistence Layer}
\label{sec:results:fixedtrace}

All preceding suites entangle agent policy with storage encoding. A scripted mock endpoint (in the artifact) replays one identical
trajectory (three reads, 182.7\,KB of observations, one answer)
through the seven persisting frameworks' unmodified adapters: the
content is fixed, so the
framework configuration is the only variable. Identical-content amplification spans
$1.03\times$ (LlamaIndex) to $6.86\times$ (InfiAgent) under the
batched transport protocol (sequential transport raises LangGraph to $7.40\times$;
Supp.\ App.~L)---a $6.7\times$
spread with no contribution from agent behavior; what remains is
the configuration-level persistence mechanism---which objects are
stored, how often, in what serialization. The protocol is deterministic, offline, and model-free
(Supp.\ Apps.~L--M).

\section{Storage in the Wild}
\label{sec:wild}

Controlled suites could exaggerate differences that vanish on
real workloads, so we compare with exported trajectories from
SWE-bench
Verified~\citep{swebench}. We harvested all 134 submissions in our
July-2026 snapshot;
keys are grouped by canonical instance ID (S3 layouts vary). Of 115
submissions with S3 prefixes, 113 expose size listings and 108 meet
the 90\%-coverage inclusion thresholds (instances and mapped bytes);
19 lack a public prefix, seven cannot be instance-normalized. Each system's trajectory volume is a
\emph{per-system lower bound} on that system's footprint---it is the
exported artifact, not the on-disk residue. Ratios between systems
inherit no such guarantee (export policies differ): the spread
corroborates heterogeneity, not runtime footprints.

\begin{figure}[t]
  \centering
  \includegraphics[width=\linewidth]{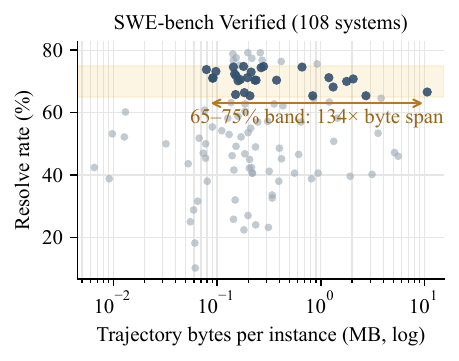}
  \caption{SWE-bench Verified, 108 instance-normalized systems:
  resolve rate vs.\ exported trajectory bytes per instance (log).
  No detectable volume--success correlation ($\tau_b{=}0.080$,
  $p{=}0.222$).}
  \label{fig:shuffle}
\end{figure}

Mean exported trajectory volume per
mapped instance ranges from 6.6\,KB to 10.6\,MB across systems---$1{,}617\times$
(Figure~\ref{fig:shuffle}). In the direct test, resolve rate shows no detectable correlation with
bytes per instance (Kendall $\tau_b{=}0.080$, permutation $p{=}0.222$):
within this population, exporting more trajectory data is not associated
with resolving more issues.
The 28 systems in the 65--75\% band alone span $134\times$ (IQR
139--402\,KB; the spread is not driven solely by extremes). Because the 108 submissions include multiple versions per team or
family (63 families; 25 multi-entry), we repeat the direct test under
two analyses that respect family-level dependence
(latest-per-family and family means; $\tau_b{\le}0.07$,
$p{\ge}0.40$)---the conclusion is unchanged. Detailed inspection of
six sampled
submissions shows intra-trajectory duplication from 1.0 (deduplicated
exports) to 3.35.

\section{A Reference Mitigation: Content-Addressed Trajectory Store}
\label{sec:cas}

Because duplication drives much of the footprint, we evaluate
content addressing---store each block once, by
hash---as a feasibility reference,
not a novel compression
algorithm. Our implementation (about 300 lines,
framework-agnostic) post-processes a run's retained stores: strings and
blobs above 1\,KB inside JSON, JSONL, log lines, and SQLite cells are
moved into a zstd-compressed content-addressed store and each is replaced
by a hash reference with preview; rehydration inverts the transform on
demand. The design preserves each framework's store
schema---composing with, not replacing, existing persistence layers.

Across 70 runs of the seven persisting frameworks, the compactor
reduces
retention by $4.8$--$32.7\times$ (Table~\ref{tab:main}, CAS column), and
all 310 retained files pass the restore checks (JSON-object equality
for structured text; row-set equality for SQLite user tables,
internal/FTS shadow tables excluded as regenerable; pass-through
byte-exact). Re-scoring
\Rep on the CAS-restored stores reproduces every framework's
original score ($\Rep{=}3$ for all six, $\Rep{=}1$ for
CrewAI)---we verify preservation rather than assume it. The compactor
achieves its largest reduction on AutoGen ($32.7\times$): its event log and per-turn snapshots repeat the
same messages---exactly what content addressing removes.

\emph{Relation to whole-store compression.} A zstd archive is
smaller than the
CAS representation (LangGraph: ${\approx}0.04$ vs.\ 0.50\,MB), but
answers a different question: \Comp measures an \emph{opaque archive},
unusable until fully decompressed, while the CAS output remains a
\emph{schema-preserving} store---blocks stay hash-addressable and
can be restored on demand (consuming references in place needs a
framework-side shim we do not build; \Rep is re-scored on restored
stores); the two compose (Supp.\ App.~H). Garbage
collection, deletion semantics, and cross-tenant privacy are out of
scope by design.

The write-task suite quantifies how far defaults are from this
floor: a 39-byte summary file (a deliberately small write task)
leaves 128\,KB--1\,MB of residue, up to $24{,}033\times$ the
artifact (the production case of \S\ref{sec:intro}, with realistic
outputs, shows ${\sim}30\times$). With the ablations, these bracket the design space:
retaining nothing yields no reconstructability, while defaults can
retain megabytes even though the same score is achievable within
50--500\,KB after post-processing.

\section{Discussion and Limitations}
\label{sec:discussion}

\emph{Isolation and backend transfer.} Containerized and sandbox
measurements differ by 0--8.5\%. Full-roster replications preserve the
storage ranking on Qwen3.6-27B ($\rho{=}0.96$; sizes 0.93--1.06$\times$)
and weak Qwen2.5-7B ($\rho{=}0.89$ despite accuracy collapse), indicating
that persistence design dominates backend choice (Table~\ref{tab:main}b;
Supp.\ App.~F).

\emph{Scope.} Six audited boundaries apply: (i) controlled suites cover
retrieval and retention, not general reasoning; (ii) results describe
versioned configurations, not frameworks in the abstract; (iii) wild data
are exports, not runtime residue; (iv) \Rep covers history only; (v) the
roster is representative and \Alpha workload-conditioned; (vi) fresh
sandboxes \emph{understate} shared threads (95.2 vs.\ 5.8\,MB; Supp.\ App.~M).

\emph{Configuration sensitivity.} AutoGen's save cadence is the most
contestable choice; an end-of-run variant remains similar (2.1 vs.\
2.49\,MB) and superlinear (\Alpha 1.50).

\Stotal and \Rep must be read jointly: either can be gamed by retaining
nothing or everything; the target is exact reconstruction at minimal
bytes (\S\ref{sec:cas}).

\emph{Why bytes matter.} The concern is less disk cost than retention
obligations: at $10^4$ tasks/day, measured defaults produce 51 versus
3.1\,GB/day for the executed history-equivalent form (Supp.\ App.~Q),
a $16\times$ gap at equal \Rep; deletion forfeits supported capabilities
(\S\ref{sec:results:ablation}).

\emph{Ethics.} Corpora are synthetic; wild data are public; CAS
obligations appear in Supp.\ App.~A.

\section{Conclusion}
\label{sec:conclusion}

Agent evaluations report task success, reliability, and inference cost,
but not persistent storage. \bench makes this axis reportable through a
serialization-aware suite, controlled and in-the-wild studies, and a
reconstruction-preserving store. Equal-accuracy configurations differ by
$15.7\times$, while the fixed-trace control isolates a $6.7\times$
persistence-layer spread. The restoration experiments further show that
exact history reconstruction need not require megabytes of retained data.
Reporting storage together with accuracy, inference cost, and
reconstructability makes the operational consequences of agent design
choices visible before deployment.

\paragraph{AI-use disclosure.} Claude (Anthropic) assisted with
parts of the experiment code and with manuscript polishing; the
authors verified and take responsibility for all content.

\bibliography{references}
\clearpage
\includepdf[pages=-]{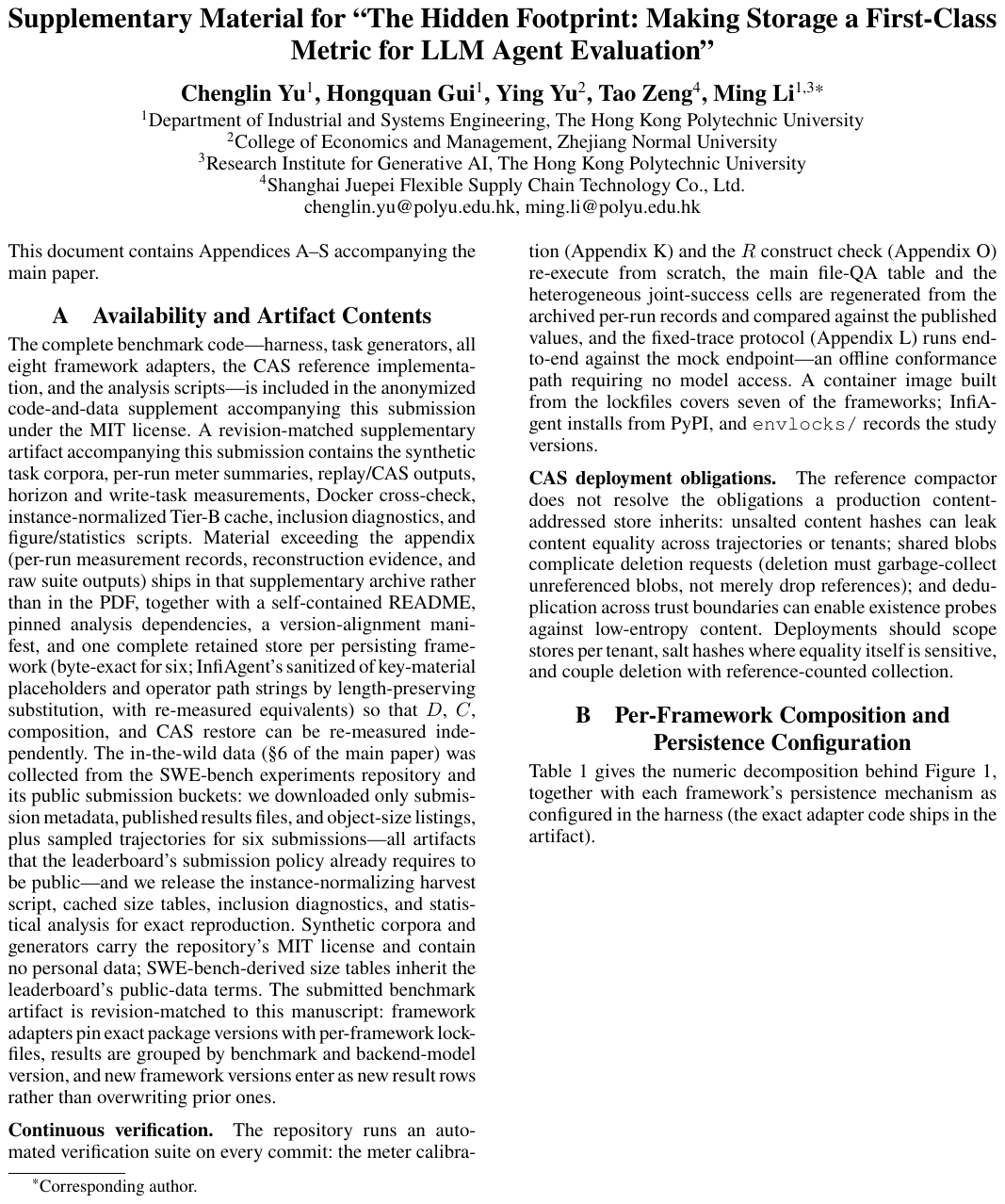}

\end{document}